%% file: main.tex
\documentclass[10pt,twocolumn,letterpaper]{article}

\usepackage{iccv}
\usepackage{times}
\usepackage{epsfig}
\usepackage{graphicx}
\usepackage{amsmath}
\usepackage{amssymb}
\usepackage{booktabs}
\usepackage[dvipsnames,table,xcdraw]{xcolor}
\usepackage[utf8]{inputenc} %
\usepackage[T1]{fontenc}    %
\usepackage{url}            %
\usepackage{amsfonts}       %
\usepackage{nicefrac}       %
\usepackage{microtype}      %
\usepackage{multirow}
\usepackage{bbm}
\usepackage{rotating}
\usepackage[english]{babel}
\usepackage{pifont}%
\usepackage[dvipsnames]{xcolor}
\usepackage[caption=false]{subfig}

\usepackage[pagebackref=true,breaklinks=true,letterpaper=true,colorlinks,bookmarks=false]{hyperref}

\input{command.tex}

\iccvfinalcopy %

\begin{document}

\newcommand{\authorsep}{\hspace{8pt}}

\title{
\vspace{-2cm} %
Hybrid Structure-from-Motion and Camera Relocalization \\
for Enhanced Egocentric Localization}

\author{Jinjie Mai$^1$ \authorsep Abdullah Hamdi$^{2}$ \authorsep Silvio Giancola$^1$ \authorsep Chen Zhao$^1$  \authorsep Bernard Ghanem$^1$\\
\normalsize$^1$King Abdullah University of Science and Technology (KAUST) \hspace{5pt} \normalsize$^2$Visual Geometry Group, University of Oxford
\\
{\tt\small \{jinjie.mai,bernard.ghanem\}@kaust.edu.sa}\\
}

\maketitle

\input{cvpr2024/1_intro}

\input{cvpr2024/acknowledgement}

{\small
\typeout{}
\bibliographystyle{ieee_fullname}
\bibliography{egbib}
}

\end{document}

%% file: command.tex
\newcommand{\mysection}[1]{\vspace{3pt}\noindent\textbf{#1.}}

\usepackage{listings}
\usepackage{color}
\definecolor{codegreen}{rgb}{0,0.6,0}
\definecolor{codegray}{rgb}{0.5,0.5,0.5}
\definecolor{codepurple}{rgb}{0.58,0,0.82}
\definecolor{backcolour}{rgb}{1,1,1}
 
\lstdefinestyle{mystyle}{
    backgroundcolor=\color{backcolour},   
    commentstyle=\color{codegreen},
    keywordstyle=\color{magenta},
    numberstyle=\tiny\color{codegray},
    stringstyle=\color{codepurple},
    basicstyle=\footnotesize,
    breakatwhitespace=false,         
    breaklines=true,                 
    captionpos=b,                    
    keepspaces=true,                 
    numbers=left,                    
    numbersep=5pt,                  
    showspaces=false,                
    showstringspaces=false,
    showtabs=false,                  
    tabsize=2
}
 

%% file: cvpr2024/1_intro.tex
\input{cvpr2024/pipeline}

\begin{abstract}
We built our pipeline EgoLoc-v1, mainly inspired by EgoLoc.
We propose a model ensemble strategy to improve the camera pose estimation part of the VQ3D task, which has been proven to be essential in previous work.
The core idea is not only to do SfM for egocentric videos but also to do 2D-3D matching between existing 3D scans and 2D video frames.
In this way, we have a hybrid SfM and camera relocalization pipeline, which can provide us with more camera poses, leading to higher QwP and overall success rate.
Our method achieves the best performance regarding the most important metric, the overall success rate. We surpass previous state-of-the-art, the competitive EgoLoc, by $1.5\%$.
The code is available at \url{https://github.com/Wayne-Mai/egoloc_v1}. 

\end{abstract}

\section{Background}

\subsection{Task definition}

Visual Queries with 3D Localization (VQ3D) is a task defined by Ego4D~\cite{grauman2022ego4d}. 
It assumes that the wearer with a head-mounted camera shows an image of the target object he wants to find at first, then the goal is to re-localize the interested object given the egocentric video stream with respect to the last frame, i.e., the relative location to the camera wearer.
Specifically, the requirement of relocalization can be formatted as predicting a 3D relative offset $\Delta d = (\Delta x, \Delta y, \Delta z)$ from the current camera's location to the groundtruth 3D  object location.

\subsection{Related works}
Ego4D~\cite{grauman2022ego4d} established a straightforward pipeline to achieve this task. They assume that they have access to the 3D Matterport scan of the scenes. Therefore, with the availability of pre-computed 3D keypoints~\cite{pc_landrieu2018large}, they extract the 2D keypoints to perform matching~\cite{sarlin20superglue}. After Perspective-n-points (PnP) and Structure-from-Motion (SfM), they can obtain the egocentric camera pose for those egocentric videos. 
Thereisnospoon~\cite{xu2022my} uses a better 2D object localization method to improve the VQ3D performance.
Eivul~\cite{mai2022estimating} is the first work that proposes to estimate camera poses independent from Matterport scans. They run COLMAP~\cite{schoenberger2016sfm} for those egocentric videos and then align these SfM camera poses with those poses from the Ego4D baseline.
In this way, they can get more camera poses, thus improving the overall success rate and final performance.
EGOCOL~\cite{forigua2023egocol} proposes to register the camera poses into the scan coordinate system regarding clips and scans. 
Most recently, EgoLoc~\cite{mai2023egoloc} gets a quite impressive performance on this task. They formalize and standardize the entire pipeline. For camera pose estimation and registration, they propose a better strategy for egocentric SfM and register the poses into the scan system by registering scan renderings into COLMAP reconstruction. For 2D object relocalization, they propose using peak detection results instead of the last track. For the final prediction, they propose to use a confidence-based weighted prediction across all visible multiview observations.

\input{cvpr2024/result_table}

\section{Method}

We built our pipeline EgoLoc-v1, mainly inspired by EgoLoc.
We propose a model ensemble strategy to improve the camera pose estimation part of the VQ3D task, which has been proven to be essential in previous work~\cite{mai2022estimating}.
Our approach is simple and straightforward, which has been illustrated in Fig.~\ref{pipeline}.
The core idea is not only to do SfM for egocentric videos but also to do 2D-3D matching between existing 3D scans and 2D video frames.
In this way, we have a hybrid SfM and camera relocalization pipeline, which can provide us with more camera poses, leading to higher QwP and overall success rate, as shown in Tab.~\ref{tb:main}.

We perform Structure from Motion (SfM)~\cite{schoenberger2016sfm} as well as EgoLoc~\cite{mai2023egoloc}, which estimates the 3D poses $\{T_0, ..., T_{N-1}\}$ for all $N$ video frames $\{k_{0},...,k_{N-1}\}$. However, in practice, we can't always obtain $N$ camera poses for all the frames because of the failure of COLMAP. To this end, we propose EgoLoc-v1, a hybrid approach of SfM and camera relocalization to get more camera poses.

Unlike EgoLoc, which only runs SfM, we additionally utilize available 3D scans to get 3D key points. 
Then we run the same feature extractor~\cite{pc_landrieu2018large} on the 2D egocentric video frames and perform matching~\cite{sarlin20superglue}. 
The proposed method is similar to the Ego4D baseline that follows the practice of classic camera relocalization methods~\cite{sarlin2019coarse}. 
We can obtain more camera poses through PnP by taking the union of the camera poses output by SfM and PnP.

After that, we follow all the other modules of EgoLoc. We obtain 2D detection results and then lift them into 3D predictions by:

\begin{align}
     [x,y,z,1]^T  = T d K^{-1}  [u,v,1]^T
    \label{eq:backprojection}
\end{align}

where $d$ is depth, $K$ is camera intrinsic, $\{u,v\}$ is the object center of detected 2D objects.

\section{Results and Conlusion}

We present our results in Tab.~\ref{tb:main}.
We compare our EgoLoc-v1 with all available methods on the public leader board. Our method achieves the best performance regarding the most important metric, the overall success rate. We surpass previous state-of-the-art, the competitive EgoLoc, by $1.5\%$.
Since the boost is marginal, it indicates that it's difficult to do matching between Matterport scans and egocentric videos, as discussed in EgoLoc~\cite{mai2023egoloc}. 

\section{Limitations}

\mysection{Speed} Since most of our components come from EgoLoc, we also suffer all the drawbacks of EgoLoc, like the slow speed of egocentric SfM.

\mysection{Scans} Unlike EgoLoc, which just requires 3D scans to register their poses for evaluation, we additionally assume there are 3D scans and key points for the recorded egocentric videos. This is a very strong assumption that may constrain the application in real cases.

%% file: cvpr2024/pipeline.tex
\begin{figure*}[!h]
    \centering
    \includegraphics[width=0.9\linewidth] {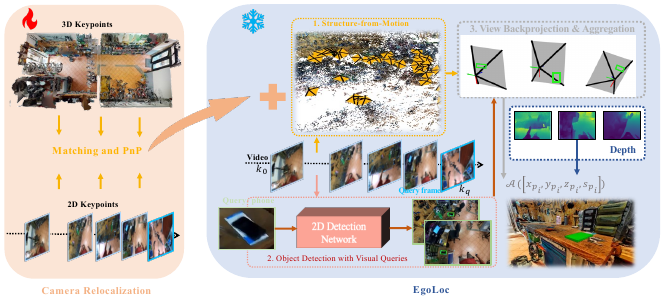}
    \caption{
        \textbf{Methodology.} 
        We propose a hybrid approach to estimate egocentric camera pose, which leads to more available prediction results for the objects of interest.
        We assume 3D keypoints are available and utilize them to perform 2D-3D matching and PnP.
        Then we combine these camera poses and those from EgoLoc SfM together to do the final prediction, i.e., the 3D position $[x,y,z]$ of the retrieved object.
    }
    \label{pipeline}
\end{figure*}

%% file: cvpr2024/result_table.tex
\begin{table}[t]
\setlength{\tabcolsep}{3pt}
\centering
\resizebox{0.99\linewidth}{!}{%
\begin{tabular}{@{}l|ccccc@{}}
\toprule
\multicolumn{1}{c}{\textbf{Method}} & \multicolumn{5}{c}{\textbf{Eval AI Test Set}} \\
 & \textbf{Succ\%↑} & \textbf{Succ*\%↑} & \textbf{L2↓} & \textbf{Angle↓} & \textbf{QwP\%↑} \\ \midrule
Ego4D~\cite{grauman2022ego4d}                & 8.71    & 51.47   & 4.93   & 1.23  & 15.15  \\
thereisnospoon~\cite{xu2022my} &9.09   & 50.60   & 4.23   &1.23  & 16.29 \\ 
Eivul~\cite{mai2022estimating}             &25.76   & 38.74   & 8.97   &1.21  & 66.29 \\ 
EGOCOL~\cite{forigua2023egocol}             &62.88   & 85.27   & 2.37   &\textbf{0.53}  & 74.62 \\ 
EgoLoc~\cite{mai2023egoloc}             &87.12   & 96.14   & 1.86   & 0.92  & 90.53 \\ 
 \textbf{EgoLoc-v1 (ours)}                 & \textbf{88.64}   & \textbf{96.15}   & \textbf{1.86}   & 1.24  & \textbf{92.05} \\ 
\bottomrule
\end{tabular}%
}
\vspace{2pt}
\caption{
        \textbf{Leaderboard performance.}
        We compare our EgoLoc-v1 with all available methods on the public leader board. Our method achieves the best performance regarding the most important metric, the overall success rate. We surpass previous state-of-the-art, the competitive EgoLoc, by $1.5\%$.
    }
    \label{tb:main}
\end{table}

%% file: cvpr2024/acknowledgement.tex
\section{Ackownledgement}
This work was supported by the KAUST Office of Sponsored Research through the GenAI Center of Excellence funding, as well as, the SDAIA-KAUST Center of Excellence in Data Science and Artificial Intelligence (SDAIA-KAUST AI).